\pdfoutput=1

\documentclass[11pt]{article}

\usepackage[preprint]{acl}

\usepackage{times}
\usepackage{latexsym}
\usepackage[T1]{fontenc}
\usepackage[utf8]{inputenc}
\usepackage{microtype}
\usepackage{inconsolata}
\usepackage{graphicx}
\usepackage{amsmath}
\usepackage[ruled,vlined,linesnumbered]{algorithm2e}
\usepackage{amssymb}
\usepackage{xcolor,soul}
\usepackage{fullpage}
\usepackage{booktabs}
\usepackage{multirow}
\usepackage{enumitem}
\usepackage[normalem]{ulem}
\useunder{\uline}{\ul}{}

\definecolor{lightblue}{RGB}{230,240,254}
\newcommand\hlc[2]{\sethlcolor{#1} \hl{#2}}

\title{Long-Form Information Alignment Evaluation Beyond Atomic Facts}
\author{Danna Zheng$^{1}$, Mirella Lapata$^{1}$, Jeff Z. Pan$^{1, 2}$ \\
            $^{1}$ School of Informatics, University of Edinburgh, UK\\
            $^{2}$ Huawei  Edinburgh Research Centre, CSI, UK\\
        dzheng@ed.ac.uk, mlap@inf.ed.ac.uk, http://knowledge-representation.org/j.z.pan/
}

\begin{document}
\maketitle

\begin{abstract}
Information alignment evaluators are vital for various NLG evaluation tasks and trustworthy LLM deployment, reducing hallucinations and enhancing user trust. Current fine-grained methods, like FactScore, verify facts individually but neglect inter-fact dependencies, enabling subtle vulnerabilities.
In this work, we introduce \textsc{MontageLie}, a challenging benchmark that constructs deceptive narratives by “montaging” truthful statements without introducing explicit hallucinations.
We demonstrate that both coarse-grained LLM-based evaluators and current fine-grained frameworks are susceptible to this attack, with AUC-ROC scores falling below 65\%.
To enable more robust fine-grained evaluation, we propose \textsc{DoveScore}, a novel framework that jointly verifies factual accuracy and event-order consistency. By modeling inter-fact relationships, \textsc{DoveScore} outperforms existing fine-grained methods by over 8\%, providing a more robust solution for long-form text alignment evaluation. 
Our code and datasets are available at \href{https://github.com/dannalily/DoveScore}{https://github.com/dannalily/DoveScore}.
\end{abstract}
\section{Introduction}

Previous studies~\cite{10.1145/3703155} have identified a critical issue in LLM deployment: factual inaccuracies, commonly known as hallucinations. To address this challenge, recent approaches~\cite{asai2024selfrag,roy-etal-2024-learning,ji-etal-2023-towards,manakul2023selfcheckgpt} introduce reflection mechanisms that perform post-hoc verification by comparing generated texts against retrieved documents or given contexts, and subsequently regenerating erroneous segments when necessary.
At the core of these approaches lies the information alignment evaluator, which determines whether a target text accurately aligns with a source text. \citet{deng-etal-2021-compression} highlight the importance of information alignment evaluation across various NLG evaluation tasks. Consequently, developing robust information alignment evaluators is essential.

Unlike fact-checking tasks \cite{si-etal-2024-checkwhy,ma-etal-2024-ex}, which typically involve short, sentence-level comparisons, information alignment evaluation often requires reasoning over extended contexts \cite{zha-etal-2023-alignscore}, where both the source and target may span multiple paragraphs. While recent advances in long-context LLMs~\cite{liu2025comprehensive} support coarse-grained alignment evaluation across entire texts, many real-world applications~\cite{min-etal-2023-factscore, zhang2024fine,ye-etal-2025-tooleyes} demand fine-grained evaluators that can assess individual factual units and pinpoint detailed inaccuracies.

\begin{figure}[t]
    \centering
    \includegraphics[width=.48\textwidth]{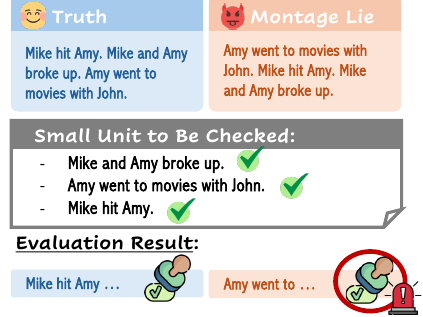}
    \caption{The figure illustrates the limitation of existing fine-grained evaluators such as FactScore and AlignScore, which struggle to detect lies composed of the exact small units that make up the truth.}
    \label{fig:limitaion-explanation}
\end{figure}

Existing fine-grained frameworks~\cite{min-etal-2023-factscore,song-etal-2024-veriscore,wei2024longform} typically decompose the target text into atomic facts, verify each against retrieved evidence, and then aggregate the results into an overall judgment. While effective at identifying surface-level errors, this approach has a critical blind spot: it overlooks relationships and dependencies between facts. Even when all individual statements are accurate, reordering them can reverse implied causal chains and mislead readers.
As illustrated in Figure~\ref{fig:limitaion-explanation}, the Lie version suggests that Amy’s outing with John triggered Mike’s violence, subtly shifting blame onto her. The Truth reveals that the violence preceded the breakup and her subsequent actions.
By altering the sequence of accurate statements, the text introduces a discourse-level manipulation that distorts causality without introducing any falsehoods. However, existing fine-grained evaluators are inherently incapable of detecting such manipulations. 

To investigate this vulnerability, we introduce \textbf{\textsc{MontageLie}}, a novel benchmark designed to test the limitations of current information alignment evaluators. Drawing inspiration from the cinematic concept of montage\footnote{https://en.wikipedia.org/wiki/Montage\_(filmmaking)}, which creates new meaning by rearranging real scenes in novel sequences, \textsc{MontageLie} constructs "montage-style lies": deceptive texts composed entirely of truthful statements, deliberately reordered to imply misleading narratives. These manipulations do not introduce fabricated facts but instead distort causal relationships by altering the sequence of events. To systematically assess model robustness, our benchmark includes four levels of difficulty, each reflecting increasing subtlety in the causal distortion. 

Such rearranging strategies, while factually accurate at the level of small textual units, exploit deceptive tactics commonly used in human communication and pose a sophisticated challenge that current evaluators are ill-equipped to detect. They also represent a realistic and underexplored attack vector in adversarial prompting~\cite{kim-etal-2024-robust} and misinformation campaigns\cite{hu2025llm,macko2025beyond}.
Experimental results demonstrate that existing fine-grained frameworks, as well as state-of-the-art long-context LLMs in coarse-grained evaluation settings, struggle to identify these subtle manipulations, achieving AUC-ROC scores consistently below 65\%.

Besides, to address the limitations of current fine-grained evaluators, we propose \textbf{\textsc{DoveScore}} (\textbf{D}escriptive and \textbf{O}rdered-Event \textbf{Ve}rification \textbf{Score}), a fine-grained evaluation framework that explicitly incorporates both atomic factual accuracy and event ordering consistency.
\textsc{DoveScore} decomposes target texts into descriptive and event-based facts, verifying their individual correctness and event sequencing against the source text, then computing a weighted precision score. Experimental evaluations show that \textsc{DoveScore} outperforms existing fine-grained methods by over 8\%.

We anticipate that both our benchmark, \textsc{MontageLie}, and our proposed evaluation method, \textsc{DoveScore}, will offer valuable insights and make a meaningful contribution to the ongoing development of robust and reliable information alignment evaluators.
\section{Background and Related Works}
\subsection{Definition of Information Alignment} 
The concept of information alignment (also termed factual consistency) was first formally defined by \citet{deng-etal-2021-compression}, who proposed a unified evaluation framework for NLG tasks and identified information alignment as a core principle. Initially, it was defined at the token level: each token in the target should be supported by the source. Later, \citet{zha-etal-2023-alignscore} introduced a more practical sequence-level definition: Text $b$ aligns with source $a$ if all information in $b$ appears accurately in $a$ without contradiction. An Information Alignment Evaluator can be formalized as:
\begin{equation}
f:(\boldsymbol{a}, \boldsymbol{b}) \rightarrow y
\end{equation}
where higher $y$ indicates stronger alignment of $b$ with respect to $a$.

\subsection{Benchmarks}
Early benchmarks mainly addressed sentence-level alignment, such as FEVER~\cite{thorne-etal-2018-fever}, FEVEROUS~\cite{aly-etal-2021-fact}, and AVERITEC~\cite{schlichtkrull2023averitec}. More recent efforts target longer texts and diverse domains. SummaC~\cite{laban-etal-2022-summac} aggregates six summarization-focused datasets, while TRUE~\cite{honovich-etal-2022-true} combines 11 datasets across summarization, dialogue, fact verification, and paraphrasing. The latest, LLM-AggreFact~\cite{tang-etal-2024-minicheck}, curates recent datasets such as AGGREFACT~\cite{tang-etal-2023-understanding}, TOFUEVAL~\cite{tang-etal-2024-tofueval}, and ClaimVerify~\cite{liu-etal-2023-evaluating}.

Despite broader coverage, these benchmarks mostly focus on unsupported or contradictory claims. Our proposed dataset, \textsc{MontageLie}, introduces a harder case: each individual claim aligns with the source, yet their combination yields a misleading narrative.

\subsection{Evaluators}
\label{sec:related-evaluators}
\paragraph{Coarse-grained Evaluators}
These methods assess alignment holistically, typically via overlap or semantic similarity. Traditional metrics like BLEU~\cite{papineni-etal-2002-bleu}, ROUGE~\cite{lin-2004-rouge}, and METEOR~\cite{banerjee-lavie-2005-meteor} emphasize lexical overlap, while embedding-based metrics (e.g., BERTScore~\cite{zhangbertscore}, BartScore~\cite{yuan2021bartscore}) and fine-tuned models (e.g., BLEURT~\cite{sellam-etal-2020-bleurt}, FactCC~\cite{kryscinski-etal-2020-evaluating}) capture semantics more effectively. However, most struggle with long-form texts due to model limitations. More recently, LLM-as-Evaluator approaches~\cite{liu2023g,luo2023chatgpt} prompt LLMs directly for quality scores, enabling more flexible, semantically rich assessments.

\paragraph{Fine-grained Evaluators}
Fine-grained methods offer interpretable, diagnostic feedback. QA-based approaches (e.g., QuestEval~\cite{scialom-etal-2021-questeval}, QAFactEval~\cite{fabbri-etal-2022-qafacteval}) extract entities, generate questions, and verify answers using the source. These methods are limited by coverage and are computationally intensive. Another strategy segments the target into sentences~\cite{laban-etal-2022-summac,zha-etal-2023-alignscore} and verifies them individually, but this ignores cross-sentence dependencies.
Recent work leverages LLMs to decompose texts into atomic facts~\cite{min-etal-2023-factscore,song-etal-2024-veriscore,song-etal-2024-finesure,wei2024longform}, offering more precise semantic units and better handling of indirect references. 

Yet, all existing methods verify claims independently, failing to detect montage lies—cases where individually correct claims, when combined, form a misleading whole. This independence assumption prevents them from capturing higher-order semantics across claims. To overcome this, we propose \textsc{DoveScore}, a novel framework that explicitly models inter-fact relationships.

\section{\textsc{MontageLie}: Information Alignment Evaluation Benchmark}
We introduce \textsc{MontageLie}, a benchmark designed to evaluate the ability of evaluators to detect misalignment in text that preserves correct individual claims but disrupts the overall intended narrative. Below we describe (1) the data construction process, (2) dataset statistics and quality checks, and (3) the evaluation metrics.

\subsection{Data Construction}
The construction process for \textsc{MontageLie} consists of three main stages: seed data sampling, montage-style lie generation, and paraphrasing. The LLM used in data construction process is \href{https://platform.openai.com/docs/models/gpt-4o-mini}{\texttt{gpt-4o-mini-2024-07-18}}, and prompts are shown in Table~\ref{tab:prompt-data-construction} in Appendix~\ref{App:PromptData}.

\subsubsection{Seed Data Sampling}
We start with publicly available long‐form summarization datasets and randomly sample pairs $(s, g)$ from each, where $s$ is the source document and $g$ is its corresponding summary, which we label as correct target text.

\begin{itemize}
  \item \textbf{SummScreen}~\cite{chen-etal-2022-summscreen}: TV-series transcripts paired with human-written recaps  capturing dialogue-driven narratives and character actions.
  \item \textbf{BookSum}~\cite{kryscinski-etal-2022-booksum}: Literary texts paired with long-form summaries, emphasizing long-range causal and temporal dependencies.
\end{itemize}

These datasets were chosen for their narrative intensity and complementary styles (dialogue versus exposition). We sample uniformly to cover a diverse range of source and target text lengths.  

\subsubsection{Montage‐Style Lie Generation}
For each correct target text $g$, we generate four montage-style lies $l_e$, $l_m$, $l_h$, and $l_{eh}$ at varying difficulty levels: easy, medium, hard, and extreme hard. 

\paragraph{Step 1: Decompose $g$ into Events $E$}
We prompt the LLM to decompose $g$ into a chronological sequence of independent events, denoted as: $E = [e_1, e_2, \dots, e_n]$. Figure~\ref{fig:dist-decomposed-event} in Appendix~\ref{App:DataDist} presents the distribution of the number of decomposed events.

\paragraph{Step 2: Shuffle $E$ with Controlled Difficulty} 
We define difficulty based on the Shuffle Degree $\operatorname{ShuffleD}$, which is to measure how out-of-order a permuted list $F = [f_1, f_2, \ldots, f_n]$ is relative to the original $E$.
Let $\pi$ be the unique permutation such that $f_i = e_{\pi(i)}$. Then $\operatorname{ShuffleD}$ is defined as:
\begin{equation}
\operatorname{ShuffleD}(E,F)=\frac{\operatorname{Inv}(\pi)}{\operatorname{Inv}_{\max }(n)} \in[0,1]
\end{equation}
where $\operatorname{Inv}(\pi)$ is the inversion count:
\begin{equation}
\scalebox{0.85}{$\displaystyle
  \operatorname{Inv}(\pi)
  = \bigl|\{(i,j)\mid 1\le i<j\le n,\;\pi(i)>\pi(j)\}\bigr|
$}
\end{equation}
and $\operatorname{Inv}_{\max}(n) = \binom{n}{2} = \frac{n(n-1)}{2}$ is the maximum possible number of inversions.

A lower value of $\operatorname{ShuffleD}$ implies a more similar sequence to the original, making the lie harder to detect. We define the difficulty levels as follows: easy when $D \in [0.80, 0.90]$, medium when $D \in [0.55, 0.65]$, hard when $D \in [0.30, 0.40]$, and extreme hard when $D \in [0.05, 0.15]$. These intervals are disjoint to ensure clear separation between difficulty levels.

To generate a permutation corresponding to a given difficulty level, we first determine the appropriate range for $\operatorname{Inv}(\pi)$ and randomly sample a target inversion count within this range. Fig~\ref{fig:dist-inversion-count} (see Appendix~\ref{App:DataDist}) shows the distribution of the sampled inversion count in our data construction process. We then use Lehmer codes~\cite{knuth1998art} to construct a permutation\footnote{Exhaustively enumerating permutations is computationally infeasible due to the combinatorial explosion of the permutation space.} that has the exact sampled inversion count. As illustrated in Algorithm~\ref{alg:random-shuffle-inversions} (see Appendix~\ref{App:ShuffleAlg}), the process begins with the maximal Lehmer code, and iteratively decrements its entries at random until the total inversion count equals the target. The resulting Lehmer code is then decoded into the corresponding permutation.

\paragraph{Step 3: Incremental Lie Generation}
After obtaining $F$, we generate the lie text incrementally. 
Giving the full $F$ to the LLM often leads to failure in preserving the order, so instead, we use an incremental generation strategy. 
We start with the first event in $F$ and sequentially add the rest. 
At each step, LLM is explicitly instructed that the new event occurs after the existing text and should be integrated as the next logical event in the narrative. The LLM is asked to continue the paragraph in a natural and coherent manner, without inserting unnecessary transitional phrases or restructuring earlier content. This approach enables precise control over event order while ensuring that the generated lie remains fluent and lexically close to the original $g$, yet semantically altered due to the reordering.

\subsubsection{Paraphrasing}
\label{sec:paraphrasing}
To test whether alignment evaluators are sensitive to narrative variation, we also generate paraphrases of both the correct target text and the lies. 
These paraphrases preserve the meaning of the original but present the events in a different narrative order or style.

We instruct LLM to rephrase the text using a different narrative technique (chronological, flashback, interjection, supplementary narration) than the original. For each correct target text $g$ and its lies $l_e$, $l_m$, $l_h$, $l_{eh}$, we generate corresponding paraphrases $g'$, $l_e'$, $l_m'$, $l_h'$, and $l_{eh}'$.

\subsection{Dataset Summary}

\subsubsection{Dataset Format and Statistic} 
Each data instance in the \textsc{MontageLie} benchmark is represented as a tuple:
\begin{equation}
    d = ⟨s, g, l_e, l_m, l_h, l_{eh}, g', l_e', l_m', l_h', l_{eh}'⟩
\end{equation}
where $s$ is the source text. The texts $g$ and $g'$ are aligned with the source text and labeled as 1, while $l_e$, $l_m$, $l_h$, $l_{eh}$ and their paraphrases are not aligned with the source and are labeled as 0.

\begin{table}[t]
\centering
\resizebox{.48\textwidth}{!}{%
\begin{tabular}{lc}
\toprule
Property & Number \\
\toprule
Total Instance & 1303 \\
Instance from BOOKSUM & 637 \\
Instance from SUMM\_SCREEN & 666 \\
Word Lengths of Source Text (Min, Max, Avg) & (312, 9937, 4201.79 ) \\
Word Lengths of  Target Text (Min, Max, Avg) & (62, 991, 258.61)\\
\bottomrule \\
\end{tabular}%
}
\caption{MontageLie Benchmark Statistic}
\label{tab:bench-statistic}
\end{table}
\begin{table}[t]
\centering
\resizebox{.48\textwidth}{!}{%
\begin{tabular}{crrrrr}
\toprule
\multirow{2}{*}{Difficulty} &\multicolumn{3}{c}{Generated Lies} & \multicolumn{2}{c}{Paraphrase} \\ \cmidrule(lr){2-4} \cmidrule(lr){5-6} 
& SemanticS & EventI & Coherence & SemanticF & StructuralV\\\midrule
Easy & 100.00 & 100.00 &94.00& 98.00& 98.00\\ 
Medium & 100.00 & 98.00& 98.00& 100.00& 98.00 \\
Hard & 98.00& 100.00& 96.00& 98.00& 100.00 \\
Extreme & 96.00&  100.00 & 98.00 & 100.00 & 98.00 \\ \bottomrule
\end{tabular}
}
\caption{Human evaluation on the quality of MontageLie Benchmark. SemanticS denotes Semantic Shift, EventI denotes Event Integrity, SemanticF denotes Semantic Fidelity, and StructuralV denotes Structural Variation.}
\label{tab:quality-check}
\end{table}

Table~\ref{tab:bench-statistic} presents detailed statistics for \textsc{MontageLie}, which comprises 1,303 data instances. Source texts contain up to 9,937 words, while target texts have lengths of up to 991 words. The length distributions are illustrated in Figure~\ref{fig:dist-bench-length} (see Appendix~\ref{App:DataDist}).

\subsubsection{Data Quality} 
To verify data quality, we conducted a human evaluation, separately assessing montage-style lies and paraphrases.

Montage-style lies were evaluated on three criteria:
\textbf{Semantic Shift} — whether the meaning differs from $g$;
\textbf{Event Integrity }— whether the core events remain unchanged (no additions, deletions, or alterations);
\textbf{Coherence} — whether the text reads smoothly, without awkward transitions or overuse of conjunctions.

Paraphrases were assessed on two criteria:
\textbf{Semantic Fidelity} — whether the meaning remains faithful to $g$;
\textbf{Structural Variation} — whether the narrative structure is meaningfully altered.

Annotators labeled each instance as \texttt{yes} or \texttt{no}. To standardize evaluation, two annotators jointly labeled five instances, achieving agreement scores of 100\%, 100\%, and 87.5\% for montage-style lies, and 87.5\% and 75.0\% for paraphrases.

We then randomly sampled 50 instances, yielding 200 lies and 200 paraphrases for human evaluation. Each annotator evaluated 25 instances.
As shown in Table~\ref{tab:quality-check}, most examples meet all criteria, confirming the overall quality of the benchmark.

\subsection{Evaluation Metrics}
We assess the effectiveness of the alignment evaluator using the AUC-ROC score, which quantifies the evaluator’s ability to distinguish the correct target texts from deceptive alternatives (“lies”) across varying levels of difficulty. 
For each target text, the evaluator assigns a score indicating how well it aligns with the given source. 
To compute the AUC-ROC for a specific difficulty level, we compare the scores of correct target texts with those of their corresponding deceptive counterparts. This process is repeated independently for each of the four difficulty levels. Finally, we report the overall effectiveness as the average AUC-ROC across all difficulty levels.

\section{\textsc{MontageLie} Challenges Evaluators}

\label{sec:exisingEval}
\begin{table*}[htbp]
\small
\centering
\resizebox{.83\textwidth}{!}{%
\begin{tabular}{cllllll}
\toprule 
\multicolumn{1}{l}{} & \multicolumn{1}{l}{Evaluator} & \multicolumn{1}{c}{Easy} & \multicolumn{1}{c}{Medium} & \multicolumn{1}{c}{Hard} & \multicolumn{1}{c}{Extreme} & \multicolumn{1}{c}{AVG} \\ \toprule
\multirow{14}{*}{Coarse-Grained} 
 & ROUGE-1 & 53.92 & 54.06 & 53.77 & 54.11 &  53.96 \\
 & ROUGE-2 & 54.57 & 54.93 & 54.71 & 54.63 & 54.71 \\
 & ROUGE-L &  53.91 & 54.15 & 53.91  & 54.40 &  54.09  \\
 & qwen-3-1.7b & 51.14 & 51.21 & 50.76 & 50.48 & 50.90 \\
 & qwen-3-4b & 60.27 & 58.90 & 57.50 &54.93  &57.90 \\
 & qwen-3-8b & 64.43 & 60.98 & 57.83 & 54.20 & 59.36\\
 & qwen-3-14b & 62.22 & 61.88 & 60.63 & 57.95  & 60.67 \\
 & qwen-3-32b & 65.80 & 65.80 & 63.13 & \textbf{59.24} & 63.49 \\ 
 & llama-3.2-instruct-1b & 50.20 &49.79 & 49.75 & 49.70 & 49.86\\ 
 & llama-3.2-instruct-3b & 49.44 &49.30 & 49.84&  50.08 & 49.66 \\
 & llama-3.1-instruct-8b &56.53 & 56.52& 55.22 &53.60 &55.47 \\
 & llama-3.3-instruct-70b &61.14 & 60.87 & 58.41& 54.44 & 58.71 \\
 & gpt-4o-mini & \textbf{68.77} & \textbf{66.39} &\textbf{63.17} & 58.57 & \textbf{64.23} \\\midrule
 \multirow{4}{*}{Fine-Grained} 
 & SummaC-ZS & 51.13 & 51.69 & 51.96 & 51.70 &  51.62 \\
 & SummaC-Conv &  \textbf{56.54} & 56.02 & 55.17 & 55.68 & 55.85 \\
 & AlignScore & 56.51 & \textbf{56.89}  &  \textbf{56.82} & \textbf{56.30} & \textbf{56.63} \\ 
 & FactScore (gpt-4o-mini) & 50.85 & 51.06  & 50.24 & 49.65 & 50.45 \\
 \bottomrule
\end{tabular}%
}
\caption{AUC-ROC Performance of existing evaluators on MontageLie.}
\label{tab:results-existing}
\end{table*}

\subsection{Evaluated Evaluators}
\paragraph{Coarse-grained Evaluators} 
\textsc{MontageLie} comprises long-form source and target texts, which limits the applicability of evaluators that are not designed to handle long contexts. Therefore, we restrict our evaluation to evaluators that support long-form input. Specifically, we report ROUGE-1, ROUGE-2, and ROUGE-L scores, as well as evaluations from LLM-as-Evaluator methods. For the latter, we adopt prompts adapted from G-Eval's consistency evaluation template (see Table~\ref{tab:prompt-data-evaluation} in Appendix~\ref{App:PromptEva}), and employ long-context LLMs, including \href{https://platform.openai.com/docs/models/gpt-4o-mini}{\texttt{gpt-4o-mini-2024-07-18}}, \href{https://huggingface.co/collections/Qwen/qwen3-67dd247413f0e2e4f653967f}{\texttt{Qwen-3}} models (1.7B, 4B, 8B, 14B, 32B), and various \href{https://huggingface.co/meta-llama}{\texttt{Llama3-instruct}} models (1B, 3B, 7B, and 70B).

\paragraph{Fine-grained Evaluators}
As discussed in Section~\ref{sec:related-evaluators}, existing fine-grained evaluators inherently struggle to detect montage-lies. To empirically validate this limitation, we report results from four representative methods: \href{https://github.com/tingofurro/summac}{SummaC-ZS}, \href{https://github.com/tingofurro/summac}{SummaC-Conv}, \href{https://github.com/yuh-zha/AlignScore}{AlignScore}, and \href{https://github.com/shmsw25/FActScore}{FactScore}. SummaC-ZS/Conv and AlignScore decompose the target text into individual sentences. SummaC-ZS/Conv leverages a natural language inference (NLI)-based model to assess the factual consistency of each sentence with the source, whereas AlignScore employs a fine-tuned model for the same purpose. In contrast, FactScore decomposes the target text into atomic facts and verifies each one via LLM prompting. In our experiments, we use \href{https://platform.openai.com/docs/models/gpt-4o-mini}{\texttt{gpt-4o-mini-2024-07-18}} as the LLM backbone for FactScore.

\subsection{Experiment Setup}
For all evaluations, we apply greedy decoding with a temperature of 0 to ensure deterministic outputs. All LLM-based evaluations are conducted in a zero-shot setting.

\subsection{Results}
\begin{figure}[t]
    \centering
    \includegraphics[width=.4\textwidth]{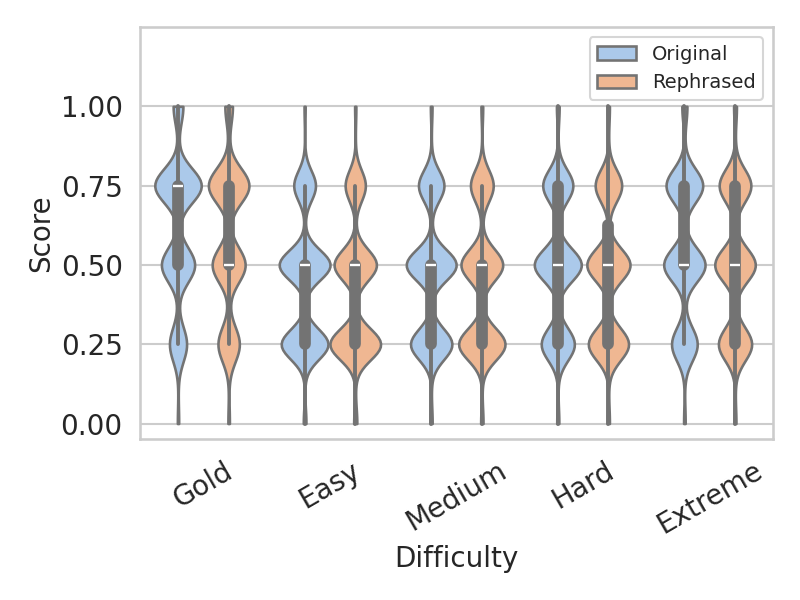}
    \caption{Violin plots of scores from \texttt{gpt-4o-mini} on \textsc{MontageLie}. The similar distributions for original and rephrased targets indicate robustness to rephrasing. Comparable trends are observed for other evaluators (see Appendix~\ref{App:ScoreDist}).}
    \label{fig:score-gpt-4o-mini}
\end{figure}
\begin{figure*}[t]
    \centering
    \includegraphics[width=\textwidth]{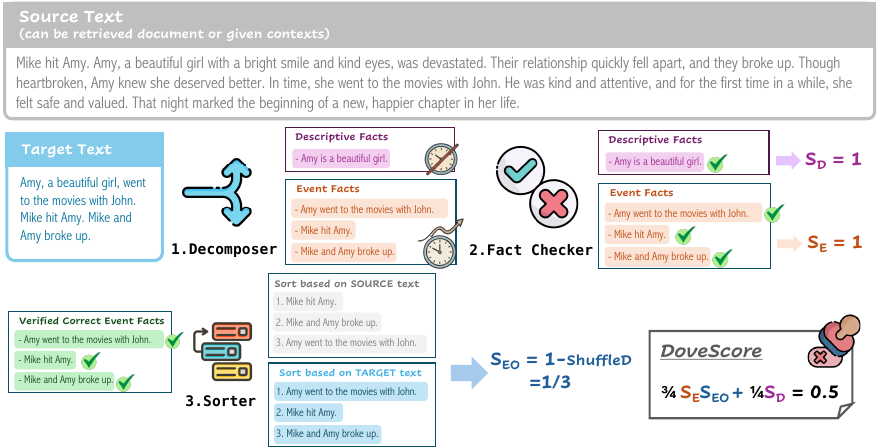}
    \caption{The illustration of DoveScore which includes three core components: the Decomposer, the Fact Checker, and the Sorter.}
    \label{fig:dove-framework}
\end{figure*}

Table~\ref{tab:results-existing} summarizes the performance of existing evaluators on the \textsc{MontageLies} benchmark.

\paragraph{Upper Bound of Current Evaluators Remains Low} 
The best-performing model, \texttt{gpt-4o-mini}, achieves an average AUC-ROC of only 64.23\%, reflecting the inherent difficulty of the task and underscoring the pressing need for more effective information alignment evaluators.

\paragraph{Fine-Grained Evaluators Suffers from Inherent Difficulty}
Fine-grained evaluation methods struggle to detect montage-style lies, as they often overlook the relationships between decomposed factual units. All evaluated methods achieve AUC-ROC scores below 57\%, highlighting their limited effectiveness. SummaC and AlignScore outperform the LLM-based FactScore, likely due to their coarser sentence-level segmentation, whereas FactScore operates at the more granular level of atomic facts. When both methods are based on \texttt{gpt-4o-mini}, the fine-grained FactScore underperforms its coarse-grained counterpart by 13.78\%.

\paragraph{Lexical Similarity Limits ROUGE's Effectiveness}
The ROUGE score achieves an AUC-ROC of approximately 54\%, as the correct target text and the montage lies are lexically similar by design during data construction. As a result, traditional n-gram-based metrics like ROUGE struggle to effectively distinguish between true and false content.

\paragraph{Small LLMs (<4B) Are Ineffective}
LLMs with fewer than 4 billion parameters in both the Qwen3 and LLaMA3 families perform poorly, achieving AUC-ROC scores below 51\%, which suggests they lack the capacity to reliably detect montage-style lies. As shown in the score distribution in Figure~\ref{fig:score-dist-coarse} (see Appendix~\ref{App:ScoreDist}), these smaller LLMs tend to assign high scores to both accurate target text and montage-style lies.

\paragraph{Qwen3 Outperform LLaMA3 Counterparts}
Across comparable model sizes, Qwen3 consistently outperforms LLaMA3: \texttt{Qwen-3-8B} outperforms \texttt{LLaMA-3.1-instruct-8B} by $\sim$4\%, and \texttt{Qwen-3-32B} surpasses \texttt{LLaMA-3.3-70B} by $\sim$5\%. 

\paragraph{Evaluators Are Robust to Narrative Variations}
We find that existing evaluators are generally robust to variations in narrative technique. As detailed in Section~\ref{sec:paraphrasing}, we rephrased each target text to alter the narrative order while preserving its original semantics. As shown in Figure~\ref{fig:score-gpt-4o-mini}, the score distributions for the original and rephrased target texts are highly similar. This suggests that current models are not easily misled by paraphrasing, a desirable property in factuality evaluation.
\section{\textsc{DoveScore}: A Fine-Grained Information Alignment Evaluation Framework}

As discussed in Section~\ref{sec:exisingEval}, current fine-grained evaluation methods are insufficiently equipped to detect  misinformation tactics like montage-style lies. To address the limitations, we propose \textsc{DoveScore}, a novel fine-grained evaluation framework designed to enable a comprehensive and nuanced assessment of information alignment.

\subsection{Method}
\begin{figure*}[htbp]
    \centering
    \includegraphics[width=\textwidth]{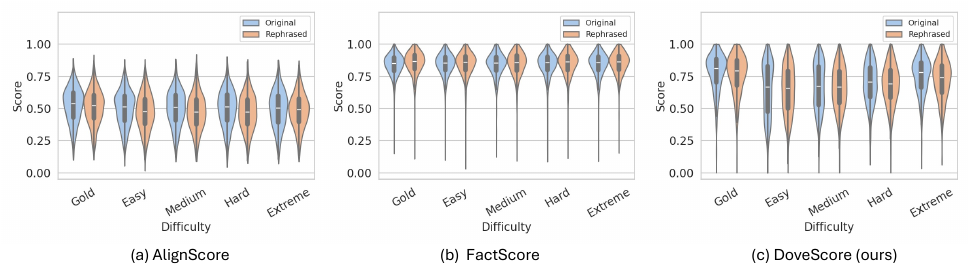}
    \caption{Score Distribution Comparison of Fine-grained Evaluators. SummaC exhibits a similar pattern to AlignScore, assigning low scores to both correct and wrong target texts (See Appendix~\ref{App:ScoreDist}).}
    \label{fig:score-dist-comp}
\end{figure*} 

As illustrated in Figure~\ref{fig:dove-framework}, the \textsc{DoveScore} framework consists of three core components: the \textit{Decomposer}, the \textit{Fact Checker}, and the \textit{Sorter}. 

\paragraph{Decomposer} 
Unlike conventional methods that uniformly segment text into sentences or atomic facts, \textsc{DoveScore} accounts for the inherent heterogeneity among factual elements. Specifically, we distinguish between two categories of facts: descriptive facts and event facts. Descriptive facts convey stable, order-independent attributes (e.g., “Octopuses have three hearts”), while event facts denote temporally ordered actions or states (e.g., “Dr. Lin submitted her resignation”). Based on this taxonomy, the decomposer partitions the target text into two lists: the event facts list ($F_E$) and the descriptive facts list ($F_D$).

\paragraph{Fact Checker}
The fact checker verifies each fact in $F_E$ and $F_D$ against the source text, resulting in two validated subsets: the set of correct event facts ($F_E^c$) and the set of correct descriptive facts ($F_D^c$). The Event Score is then computed as $S_E = \frac{|F_E^c|}{|F_E|}$, and the Descriptive Score as $S_D = \frac{|F_D^c|}{|F_D|}$.

\paragraph{Sorter} 
The sorter reorganizes the verified correct event facts $F_E^c$ into two ordered sequences. The first sequence, denoted $\operatorname{Sorted}(F_E^c, s)$, reflects their chronological order in the source text, while the second, $\operatorname{Sorted}(F_E^c, t)$, reflects their order in the target text. The similarity between these sequences is measured by the Event Order Score, defined as $S_{EO} = 1 - \operatorname{ShuffleD}\big(\operatorname{Sorted}(F_E^c, s), \operatorname{Sorted}(F_E^c, t)\big)$.

\paragraph{Score Computation} 
 The final DoveScore is computed by combining $S_E$,  $S_{EO}$,  $S_D$ using a frequency-based weighting factor $\alpha = \frac{|F_E|}{|F_E| + |F_D|}$, which adjusts the relative importance of event and descriptive facts according to their frequency in the target text:

\begin{equation}
\text{DoveScore} = \alpha \cdot S_E \cdot S_{EO} + (1 - \alpha) \cdot S_D
\end{equation}

\subsection{Experiment Setup}
For the decomposer, fact checker, and sorter used in this experiment, we prompt LLM with the prompts as shown in Table~\ref{tab:prompt-dovescore-decomposer}, Table~\ref{tab:prompt-dovescore-fact-checker} and Table~\ref{tab:prompt-dovescore-sorter}. We exampled with \href{https://platform.openai.com/docs/models/gpt-4o-mini}{\texttt{gpt-4o-mini-2024-07-18}} as the LLM with temperature 0.

\subsection{Results}
\begin{table}[t]
\centering
\resizebox{.48\textwidth}{!}{%
\begin{tabular}{lccccc}
\toprule
 & Easy & Medium & Hard & Extreme & AVG \\ \toprule
Coarse-Grained & {\ul 68.77} & {\ul 66.39} & {\ul 63.17} & \textbf{59.24} & {\ul 64.23} \\
Fine-Grained & 56.54 & 56.89 & 56.82 & 56.30 & 56.63 \\ \midrule
\begin{tabular}[c]{@{}l@{}}DoveScore \\ (Fine-Grained)\end{tabular} & \textbf{69.06} & \textbf{68.27} & \textbf{65.80} & {\ul 57.87} & \textbf{65.25} \\ \bottomrule
\end{tabular}%
}
\caption{ROC-AUC of \textsc{DoveScore} on \textsc{MontageLie}, compared to top scores by existing evaluators at each difficulty level. \textbf{Bold} marks the best, \underline{underlined} the second best.}
\label{tab:result-dovescore}
\end{table}
As shown in Table~\ref{tab:results-existing}, \textsc{DoveScore} achieves the highest average AUC-ROC score of 65.25\%, outperforming existing fine-grained evaluators by over 8\%. Compared to FactScore, which uses the same LLM backbone, \textsc{DoveScore} improves performance by 14.8\%. The distribution of its sub-scores ($S_E$, $S_{EO}$, $S_D$) in Figure~\ref{fig:score-dist-subdove} (Appendix~\ref{App:ScoreDist}) highlights the contribution of $S_{EO}$ in enhancing the model’s ability to distinguish between truthful and deceptive texts.

Further evidence from Figure~\ref{fig:score-dist-comp} reveals systematic differences in how evaluators handle complex deception styles. Sentence-split-based methods such as SummaC and AlignScore show limited discrimination, often assigning similarly low scores to both correct and incorrect targets—likely due to inference ambiguity from rigid segmentation. FactScore, which evaluates at the fact level, tends to assign uniformly high scores across targets, ignoring inter-fact coherence. In contrast, \textsc{DoveScore} consistently assigns higher scores to correct targets and lower scores to deceptive ones, reflecting stronger discrimination capabilities and robustness to diverse misinformation strategies.

\section{Conclusion}

In this work, we present \textsc{MontageLie}, a novel benchmark designed to reveal a critical vulnerability in current information alignment evaluators: their inability to detect misleading narratives composed of reordered yet truthful statements. We show that both coarse-grained and fine-grained methods struggle with such manipulations, with AUC-ROC scores falling below 65\% on \textsc{MontageLie}. 
We propose \textsc{DoveScore}, a fine-grained evaluation framework that jointly considers factual accuracy and event order consistency, improving performance from 50.45\% to 65.25\% over FactScore.
\textsc{DoveScore} is designed as a modular framework, allowing each component to be independently refined and improved. Among these, the sorter stands out as a critical and currently underexplored component that deserves targeted research efforts.
While substantial room for further exploration remains, our work marks an important step toward more robust alignment evaluation for long-form content.

\section*{Limitations}
\paragraph{MontageLie Benchmark} The MontageLie benchmark used in our study is entirely generated by large language models rather than written by human annotators. While this approach enables scalable and diverse data generation, it may introduce distributional artifacts or stylistic patterns that do not fully reflect real-world human-written misinformation. Moreover, the benchmark currently only covers English. Extending MontageLie to include human-curated data and multilingual variants would improve its generalizability and practical relevance.
\paragraph{DoveScore Framework} In this work, we demonstrate DoveScore using GPT-4o-mini as the backbone model. However, DoveScore is designed as a modular and model-agnostic framework—each component (e.g., evidence extractor, fact scorer, and sorter) can be flexibly instantiated with different language models. Future work could explore alternative backbones to assess the robustness and adaptability of the framework under varying resource constraints and capabilities. Within DoveScore, our current sorter module takes the full list of candidate responses as input and predicts a globally reordered list. This design is chosen over pairwise comparison-based sorting methods to reduce computational complexity. However, the trade-off between efficiency and ranking accuracy remains an open research question. More sophisticated or hybrid sorting strategies may offer better performance while maintaining tractable runtime.

\section*{Ethics Statement}
All data used in this study are derived from publicly available datasets and do not contain any personally identifiable or sensitive information. The additional data used for the MontageLie benchmark were generated using LLMs. To ensure the quality of the generated data, we conducted a manual evaluation with two human annotators: one is one of the authors of this paper, and the other is an external contributor who received compensation at the standard hourly rate designated for tutors and demonstrators at our university.


\bibliography{custom}

\begin{thebibliography}{42}
\providecommand{\natexlab}[1]{#1}

\bibitem[{Aly et~al.(2021)Aly, Guo, Schlichtkrull, Thorne, Vlachos, Christodoulopoulos, Cocarascu, and Mittal}]{aly-etal-2021-fact}
Rami Aly, Zhijiang Guo, Michael~Sejr Schlichtkrull, James Thorne, Andreas Vlachos, Christos Christodoulopoulos, Oana Cocarascu, and Arpit Mittal. 2021.
\newblock \href {https://doi.org/10.18653/v1/2021.fever-1.1} {The fact extraction and {VER}ification over unstructured and structured information ({FEVEROUS}) shared task}.
\newblock In \emph{Proceedings of the Fourth Workshop on Fact Extraction and VERification (FEVER)}, pages 1--13, Dominican Republic. Association for Computational Linguistics.

\bibitem[{Asai et~al.(2024)Asai, Wu, Wang, Sil, and Hajishirzi}]{asai2024selfrag}
Akari Asai, Zeqiu Wu, Yizhong Wang, Avirup Sil, and Hannaneh Hajishirzi. 2024.
\newblock \href {https://openreview.net/forum?id=hSyW5go0v8} {Self-{RAG}: Learning to retrieve, generate, and critique through self-reflection}.
\newblock In \emph{The Twelfth International Conference on Learning Representations}.

\bibitem[{Banerjee and Lavie(2005)}]{banerjee-lavie-2005-meteor}
Satanjeev Banerjee and Alon Lavie. 2005.
\newblock \href {https://aclanthology.org/W05-0909/} {{METEOR}: An automatic metric for {MT} evaluation with improved correlation with human judgments}.
\newblock In \emph{Proceedings of the {ACL} Workshop on Intrinsic and Extrinsic Evaluation Measures for Machine Translation and/or Summarization}, pages 65--72, Ann Arbor, Michigan. Association for Computational Linguistics.

\bibitem[{Chen et~al.(2022)Chen, Chu, Wiseman, and Gimpel}]{chen-etal-2022-summscreen}
Mingda Chen, Zewei Chu, Sam Wiseman, and Kevin Gimpel. 2022.
\newblock \href {https://doi.org/10.18653/v1/2022.acl-long.589} {{S}umm{S}creen: A dataset for abstractive screenplay summarization}.
\newblock In \emph{Proceedings of the 60th Annual Meeting of the Association for Computational Linguistics (Volume 1: Long Papers)}, pages 8602--8615, Dublin, Ireland. Association for Computational Linguistics.

\bibitem[{Deng et~al.(2021)Deng, Tan, Liu, Xing, and Hu}]{deng-etal-2021-compression}
Mingkai Deng, Bowen Tan, Zhengzhong Liu, Eric Xing, and Zhiting Hu. 2021.
\newblock \href {https://doi.org/10.18653/v1/2021.emnlp-main.599} {Compression, transduction, and creation: A unified framework for evaluating natural language generation}.
\newblock In \emph{Proceedings of the 2021 Conference on Empirical Methods in Natural Language Processing}, pages 7580--7605, Online and Punta Cana, Dominican Republic. Association for Computational Linguistics.

\bibitem[{Fabbri et~al.(2022)Fabbri, Wu, Liu, and Xiong}]{fabbri-etal-2022-qafacteval}
Alexander Fabbri, Chien-Sheng Wu, Wenhao Liu, and Caiming Xiong. 2022.
\newblock \href {https://doi.org/10.18653/v1/2022.naacl-main.187} {{QAF}act{E}val: Improved {QA}-based factual consistency evaluation for summarization}.
\newblock In \emph{Proceedings of the 2022 Conference of the North American Chapter of the Association for Computational Linguistics: Human Language Technologies}, pages 2587--2601, Seattle, United States. Association for Computational Linguistics.

\bibitem[{Honovich et~al.(2022)Honovich, Aharoni, Herzig, Taitelbaum, Kukliansy, Cohen, Scialom, Szpektor, Hassidim, and Matias}]{honovich-etal-2022-true}
Or~Honovich, Roee Aharoni, Jonathan Herzig, Hagai Taitelbaum, Doron Kukliansy, Vered Cohen, Thomas Scialom, Idan Szpektor, Avinatan Hassidim, and Yossi Matias. 2022.
\newblock \href {https://doi.org/10.18653/v1/2022.dialdoc-1.19} {{TRUE}: Re-evaluating factual consistency evaluation}.
\newblock In \emph{Proceedings of the Second DialDoc Workshop on Document-grounded Dialogue and Conversational Question Answering}, pages 161--175, Dublin, Ireland. Association for Computational Linguistics.

\bibitem[{Hu et~al.(2025)Hu, Sheng, Cao, Li, and Wang}]{hu2025llm}
Beizhe Hu, Qiang Sheng, Juan Cao, Yang Li, and Danding Wang. 2025.
\newblock Llm-generated fake news induces truth decay in news ecosystem: A case study on neural news recommendation.
\newblock \emph{arXiv preprint arXiv:2504.20013}.

\bibitem[{Huang et~al.(2025)Huang, Yu, Ma, Zhong, Feng, Wang, Chen, Peng, Feng, Qin, and Liu}]{10.1145/3703155}
Lei Huang, Weijiang Yu, Weitao Ma, Weihong Zhong, Zhangyin Feng, Haotian Wang, Qianglong Chen, Weihua Peng, Xiaocheng Feng, Bing Qin, and Ting Liu. 2025.
\newblock \href {https://doi.org/10.1145/3703155} {A survey on hallucination in large language models: Principles, taxonomy, challenges, and open questions}.
\newblock \emph{ACM Trans. Inf. Syst.}, 43(2).

\bibitem[{Ji et~al.(2023)Ji, Yu, Xu, Lee, Ishii, and Fung}]{ji-etal-2023-towards}
Ziwei Ji, Tiezheng Yu, Yan Xu, Nayeon Lee, Etsuko Ishii, and Pascale Fung. 2023.
\newblock \href {https://doi.org/10.18653/v1/2023.findings-emnlp.123} {Towards mitigating {LLM} hallucination via self reflection}.
\newblock In \emph{Findings of the Association for Computational Linguistics: EMNLP 2023}, pages 1827--1843, Singapore. Association for Computational Linguistics.

\bibitem[{Kim et~al.(2024)Kim, Derakhshan, and Harris}]{kim-etal-2024-robust}
Jinhwa Kim, Ali Derakhshan, and Ian Harris. 2024.
\newblock \href {https://doi.org/10.18653/v1/2024.woah-1.12} {Robust safety classifier against jailbreaking attacks: Adversarial prompt shield}.
\newblock In \emph{Proceedings of the 8th Workshop on Online Abuse and Harms (WOAH 2024)}, pages 159--170, Mexico City, Mexico. Association for Computational Linguistics.

\bibitem[{Knuth(1998)}]{knuth1998art}
Donald~E Knuth. 1998.
\newblock \emph{The Art of Computer Programming: Sorting and Searching, volume 3}.
\newblock Addison-Wesley Professional.

\bibitem[{Kryscinski et~al.(2020)Kryscinski, McCann, Xiong, and Socher}]{kryscinski-etal-2020-evaluating}
Wojciech Kryscinski, Bryan McCann, Caiming Xiong, and Richard Socher. 2020.
\newblock \href {https://doi.org/10.18653/v1/2020.emnlp-main.750} {Evaluating the factual consistency of abstractive text summarization}.
\newblock In \emph{Proceedings of the 2020 Conference on Empirical Methods in Natural Language Processing (EMNLP)}, pages 9332--9346, Online. Association for Computational Linguistics.

\bibitem[{Kryscinski et~al.(2022)Kryscinski, Rajani, Agarwal, Xiong, and Radev}]{kryscinski-etal-2022-booksum}
Wojciech Kryscinski, Nazneen Rajani, Divyansh Agarwal, Caiming Xiong, and Dragomir Radev. 2022.
\newblock \href {https://doi.org/10.18653/v1/2022.findings-emnlp.488} {{BOOKSUM}: A collection of datasets for long-form narrative summarization}.
\newblock In \emph{Findings of the Association for Computational Linguistics: EMNLP 2022}, pages 6536--6558, Abu Dhabi, United Arab Emirates. Association for Computational Linguistics.

\bibitem[{Laban et~al.(2022)Laban, Schnabel, Bennett, and Hearst}]{laban-etal-2022-summac}
Philippe Laban, Tobias Schnabel, Paul~N. Bennett, and Marti~A. Hearst. 2022.
\newblock \href {https://doi.org/10.1162/tacl_a_00453} {{S}umma{C}: Re-visiting {NLI}-based models for inconsistency detection in summarization}.
\newblock \emph{Transactions of the Association for Computational Linguistics}, 10:163--177.

\bibitem[{Lin(2004)}]{lin-2004-rouge}
Chin-Yew Lin. 2004.
\newblock \href {https://aclanthology.org/W04-1013/} {{ROUGE}: A package for automatic evaluation of summaries}.
\newblock In \emph{Text Summarization Branches Out}, pages 74--81, Barcelona, Spain. Association for Computational Linguistics.

\bibitem[{Liu et~al.(2025)Liu, Zhu, Bai, He, Liao, Que, Wang, Zhang, Zhang, Zhang et~al.}]{liu2025comprehensive}
Jiaheng Liu, Dawei Zhu, Zhiqi Bai, Yancheng He, Huanxuan Liao, Haoran Que, Zekun Wang, Chenchen Zhang, Ge~Zhang, Jiebin Zhang, and 1 others. 2025.
\newblock A comprehensive survey on long context language modeling.
\newblock \emph{arXiv preprint arXiv:2503.17407}.

\bibitem[{Liu et~al.(2023{\natexlab{a}})Liu, Zhang, and Liang}]{liu-etal-2023-evaluating}
Nelson Liu, Tianyi Zhang, and Percy Liang. 2023{\natexlab{a}}.
\newblock \href {https://doi.org/10.18653/v1/2023.findings-emnlp.467} {Evaluating verifiability in generative search engines}.
\newblock In \emph{Findings of the Association for Computational Linguistics: EMNLP 2023}, pages 7001--7025, Singapore. Association for Computational Linguistics.

\bibitem[{Liu et~al.(2023{\natexlab{b}})Liu, Iter, Xu, Wang, Xu, and Zhu}]{liu2023g}
Yang Liu, Dan Iter, Yichong Xu, Shuohang Wang, Ruochen Xu, and Chenguang Zhu. 2023{\natexlab{b}}.
\newblock G-eval: Nlg evaluation using gpt-4 with better human alignment.
\newblock In \emph{Proceedings of the 2023 Conference on Empirical Methods in Natural Language Processing}, pages 2511--2522.

\bibitem[{Luo et~al.(2023)Luo, Xie, and Ananiadou}]{luo2023chatgpt}
Zheheng Luo, Qianqian Xie, and Sophia Ananiadou. 2023.
\newblock Chatgpt as a factual inconsistency evaluator for text summarization.
\newblock \emph{arXiv preprint arXiv:2303.15621}.

\bibitem[{Ma et~al.(2024)Ma, Xu, Wei, Chen, Wang, Liu, Wu, and Wang}]{ma-etal-2024-ex}
Huanhuan Ma, Weizhi Xu, Yifan Wei, Liuji Chen, Liang Wang, Qiang Liu, Shu Wu, and Liang Wang. 2024.
\newblock \href {https://doi.org/10.18653/v1/2024.findings-acl.556} {{EX}-{FEVER}: A dataset for multi-hop explainable fact verification}.
\newblock In \emph{Findings of the Association for Computational Linguistics: ACL 2024}, pages 9340--9353, Bangkok, Thailand. Association for Computational Linguistics.

\bibitem[{Macko et~al.(2025)Macko, Ramakrishnan, Lucas, Moro, Srba, Uchendu, and Lee}]{macko2025beyond}
Dominik Macko, Aashish~Anantha Ramakrishnan, Jason~Samuel Lucas, Robert Moro, Ivan Srba, Adaku Uchendu, and Dongwon Lee. 2025.
\newblock Beyond speculation: Measuring the growing presence of llm-generated texts in multilingual disinformation.
\newblock \emph{arXiv preprint arXiv:2503.23242}.

\bibitem[{Manakul et~al.(2023)Manakul, Liusie, and Gales}]{manakul2023selfcheckgpt}
Potsawee Manakul, Adian Liusie, and Mark Gales. 2023.
\newblock \href {https://openreview.net/forum?id=RwzFNbJ3Ez} {Selfcheck{GPT}: Zero-resource black-box hallucination detection for generative large language models}.
\newblock In \emph{The 2023 Conference on Empirical Methods in Natural Language Processing}.

\bibitem[{Min et~al.(2023)Min, Krishna, Lyu, Lewis, Yih, Koh, Iyyer, Zettlemoyer, and Hajishirzi}]{min-etal-2023-factscore}
Sewon Min, Kalpesh Krishna, Xinxi Lyu, Mike Lewis, Wen-tau Yih, Pang Koh, Mohit Iyyer, Luke Zettlemoyer, and Hannaneh Hajishirzi. 2023.
\newblock \href {https://doi.org/10.18653/v1/2023.emnlp-main.741} {{FA}ct{S}core: Fine-grained atomic evaluation of factual precision in long form text generation}.
\newblock In \emph{Proceedings of the 2023 Conference on Empirical Methods in Natural Language Processing}, pages 12076--12100, Singapore. Association for Computational Linguistics.

\bibitem[{Papineni et~al.(2002)Papineni, Roukos, Ward, and Zhu}]{papineni-etal-2002-bleu}
Kishore Papineni, Salim Roukos, Todd Ward, and Wei-Jing Zhu. 2002.
\newblock \href {https://doi.org/10.3115/1073083.1073135} {{B}leu: a method for automatic evaluation of machine translation}.
\newblock In \emph{Proceedings of the 40th Annual Meeting of the Association for Computational Linguistics}, pages 311--318, Philadelphia, Pennsylvania, USA. Association for Computational Linguistics.

\bibitem[{Roy et~al.(2024)Roy, Ribeiro, Blloshmi, and Small}]{roy-etal-2024-learning}
Nirmal Roy, Leonardo F.~R. Ribeiro, Rexhina Blloshmi, and Kevin Small. 2024.
\newblock \href {https://doi.org/10.18653/v1/2024.findings-emnlp.622} {Learning when to retrieve, what to rewrite, and how to respond in conversational {QA}}.
\newblock In \emph{Findings of the Association for Computational Linguistics: EMNLP 2024}, pages 10604--10625, Miami, Florida, USA. Association for Computational Linguistics.

\bibitem[{Schlichtkrull et~al.(2023)Schlichtkrull, Guo, and Vlachos}]{schlichtkrull2023averitec}
Michael~Sejr Schlichtkrull, Zhijiang Guo, and Andreas Vlachos. 2023.
\newblock \href {https://openreview.net/forum?id=fKzSz0oyaI} {{AV}eritec: A dataset for real-world claim verification with evidence from the web}.
\newblock In \emph{Thirty-seventh Conference on Neural Information Processing Systems Datasets and Benchmarks Track}.

\bibitem[{Scialom et~al.(2021)Scialom, Dray, Lamprier, Piwowarski, Staiano, Wang, and Gallinari}]{scialom-etal-2021-questeval}
Thomas Scialom, Paul-Alexis Dray, Sylvain Lamprier, Benjamin Piwowarski, Jacopo Staiano, Alex Wang, and Patrick Gallinari. 2021.
\newblock \href {https://doi.org/10.18653/v1/2021.emnlp-main.529} {{Q}uest{E}val: Summarization asks for fact-based evaluation}.
\newblock In \emph{Proceedings of the 2021 Conference on Empirical Methods in Natural Language Processing}, pages 6594--6604, Online and Punta Cana, Dominican Republic. Association for Computational Linguistics.

\bibitem[{Sellam et~al.(2020)Sellam, Das, and Parikh}]{sellam-etal-2020-bleurt}
Thibault Sellam, Dipanjan Das, and Ankur Parikh. 2020.
\newblock \href {https://doi.org/10.18653/v1/2020.acl-main.704} {{BLEURT}: Learning robust metrics for text generation}.
\newblock In \emph{Proceedings of the 58th Annual Meeting of the Association for Computational Linguistics}, pages 7881--7892, Online. Association for Computational Linguistics.

\bibitem[{Si et~al.(2024)Si, Zhao, Zhu, Zhu, Lu, and Zhou}]{si-etal-2024-checkwhy}
Jiasheng Si, Yibo Zhao, Yingjie Zhu, Haiyang Zhu, Wenpeng Lu, and Deyu Zhou. 2024.
\newblock \href {https://doi.org/10.18653/v1/2024.acl-long.835} {{CHECKWHY}: Causal fact verification via argument structure}.
\newblock In \emph{Proceedings of the 62nd Annual Meeting of the Association for Computational Linguistics (Volume 1: Long Papers)}, pages 15636--15659, Bangkok, Thailand. Association for Computational Linguistics.

\bibitem[{Song et~al.(2024{\natexlab{a}})Song, Su, Shalyminov, Cai, and Mansour}]{song-etal-2024-finesure}
Hwanjun Song, Hang Su, Igor Shalyminov, Jason Cai, and Saab Mansour. 2024{\natexlab{a}}.
\newblock \href {https://doi.org/10.18653/v1/2024.acl-long.51} {{F}ine{S}ur{E}: Fine-grained summarization evaluation using {LLM}s}.
\newblock In \emph{Proceedings of the 62nd Annual Meeting of the Association for Computational Linguistics (Volume 1: Long Papers)}, pages 906--922, Bangkok, Thailand. Association for Computational Linguistics.

\bibitem[{Song et~al.(2024{\natexlab{b}})Song, Kim, and Iyyer}]{song-etal-2024-veriscore}
Yixiao Song, Yekyung Kim, and Mohit Iyyer. 2024{\natexlab{b}}.
\newblock \href {https://doi.org/10.18653/v1/2024.findings-emnlp.552} {{V}eri{S}core: Evaluating the factuality of verifiable claims in long-form text generation}.
\newblock In \emph{Findings of the Association for Computational Linguistics: EMNLP 2024}, pages 9447--9474, Miami, Florida, USA. Association for Computational Linguistics.

\bibitem[{Tang et~al.(2023)Tang, Goyal, Fabbri, Laban, Xu, Yavuz, Kryscinski, Rousseau, and Durrett}]{tang-etal-2023-understanding}
Liyan Tang, Tanya Goyal, Alex Fabbri, Philippe Laban, Jiacheng Xu, Semih Yavuz, Wojciech Kryscinski, Justin Rousseau, and Greg Durrett. 2023.
\newblock \href {https://doi.org/10.18653/v1/2023.acl-long.650} {Understanding factual errors in summarization: Errors, summarizers, datasets, error detectors}.
\newblock In \emph{Proceedings of the 61st Annual Meeting of the Association for Computational Linguistics (Volume 1: Long Papers)}, pages 11626--11644, Toronto, Canada. Association for Computational Linguistics.

\bibitem[{Tang et~al.(2024{\natexlab{a}})Tang, Laban, and Durrett}]{tang-etal-2024-minicheck}
Liyan Tang, Philippe Laban, and Greg Durrett. 2024{\natexlab{a}}.
\newblock \href {https://doi.org/10.18653/v1/2024.emnlp-main.499} {{M}ini{C}heck: Efficient fact-checking of {LLM}s on grounding documents}.
\newblock In \emph{Proceedings of the 2024 Conference on Empirical Methods in Natural Language Processing}, pages 8818--8847, Miami, Florida, USA. Association for Computational Linguistics.

\bibitem[{Tang et~al.(2024{\natexlab{b}})Tang, Shalyminov, Wong, Burnsky, Vincent, Yang, Singh, Feng, Song, Su, Sun, Zhang, Mansour, and McKeown}]{tang-etal-2024-tofueval}
Liyan Tang, Igor Shalyminov, Amy Wong, Jon Burnsky, Jake Vincent, Yu{'}an Yang, Siffi Singh, Song Feng, Hwanjun Song, Hang Su, Lijia Sun, Yi~Zhang, Saab Mansour, and Kathleen McKeown. 2024{\natexlab{b}}.
\newblock \href {https://doi.org/10.18653/v1/2024.naacl-long.251} {{T}ofu{E}val: Evaluating hallucinations of {LLM}s on topic-focused dialogue summarization}.
\newblock In \emph{Proceedings of the 2024 Conference of the North American Chapter of the Association for Computational Linguistics: Human Language Technologies (Volume 1: Long Papers)}, pages 4455--4480, Mexico City, Mexico. Association for Computational Linguistics.

\bibitem[{Thorne et~al.(2018)Thorne, Vlachos, Christodoulopoulos, and Mittal}]{thorne-etal-2018-fever}
James Thorne, Andreas Vlachos, Christos Christodoulopoulos, and Arpit Mittal. 2018.
\newblock \href {https://doi.org/10.18653/v1/N18-1074} {{FEVER}: a large-scale dataset for fact extraction and {VER}ification}.
\newblock In \emph{Proceedings of the 2018 Conference of the North {A}merican Chapter of the Association for Computational Linguistics: Human Language Technologies, Volume 1 (Long Papers)}, pages 809--819, New Orleans, Louisiana. Association for Computational Linguistics.

\bibitem[{Wei et~al.(2024)Wei, Yang, Song, Lu, Hu, Huang, Tran, Peng, Liu, Huang, Du, and Le}]{wei2024longform}
Jerry Wei, Chengrun Yang, Xinying Song, Yifeng Lu, Nathan~Zixia Hu, Jie Huang, Dustin Tran, Daiyi Peng, Ruibo Liu, Da~Huang, Cosmo Du, and Quoc~V Le. 2024.
\newblock \href {https://openreview.net/forum?id=4M9f8VMt2C} {Long-form factuality in large language models}.
\newblock In \emph{The Thirty-eighth Annual Conference on Neural Information Processing Systems}.

\bibitem[{Ye et~al.(2025)Ye, Li, Gao, Huang, Wu, Li, Fan, Dou, Ji, Zhang, Gui, and Huang}]{ye-etal-2025-tooleyes}
Junjie Ye, Guanyu Li, SongYang Gao, Caishuang Huang, Yilong Wu, Sixian Li, Xiaoran Fan, Shihan Dou, Tao Ji, Qi~Zhang, Tao Gui, and Xuanjing Huang. 2025.
\newblock \href {https://aclanthology.org/2025.coling-main.12/} {{T}ool{E}yes: Fine-grained evaluation for tool learning capabilities of large language models in real-world scenarios}.
\newblock In \emph{Proceedings of the 31st International Conference on Computational Linguistics}, pages 156--187, Abu Dhabi, UAE. Association for Computational Linguistics.

\bibitem[{Yuan et~al.(2021)Yuan, Neubig, and Liu}]{yuan2021bartscore}
Weizhe Yuan, Graham Neubig, and Pengfei Liu. 2021.
\newblock \href {https://openreview.net/forum?id=5Ya8PbvpZ9} {{BARTS}core: Evaluating generated text as text generation}.
\newblock In \emph{Advances in Neural Information Processing Systems}.

\bibitem[{Zha et~al.(2023)Zha, Yang, Li, and Hu}]{zha-etal-2023-alignscore}
Yuheng Zha, Yichi Yang, Ruichen Li, and Zhiting Hu. 2023.
\newblock \href {https://doi.org/10.18653/v1/2023.acl-long.634} {{A}lign{S}core: Evaluating factual consistency with a unified alignment function}.
\newblock In \emph{Proceedings of the 61st Annual Meeting of the Association for Computational Linguistics (Volume 1: Long Papers)}, pages 11328--11348, Toronto, Canada. Association for Computational Linguistics.

\bibitem[{Zhang et~al.(2020)Zhang, Kishore, Wu, Weinberger, and Artzi}]{zhangbertscore}
Tianyi Zhang, Varsha Kishore, Felix Wu, Kilian~Q Weinberger, and Yoav Artzi. 2020.
\newblock Bertscore: Evaluating text generation with bert.
\newblock In \emph{International Conference on Learning Representations}.

\bibitem[{Zhang et~al.(2024)Zhang, Zuo, and Jing}]{zhang2024fine}
Yue Zhang, Jingxuan Zuo, and Liqiang Jing. 2024.
\newblock Fine-grained and explainable factuality evaluation for multimodal summarization.
\newblock \emph{arXiv preprint arXiv:2402.11414}.

\end{thebibliography}

\appendix
\section{Prompts}
\subsection{Prompts Used in Data Construction}
\label{App:PromptData}
\begin{table*}[ht]
    \centering
    \small
    \noindent\fbox{%
    \begin{minipage}{2.0\columnwidth} 
\ttfamily
\textbf{\underline{Prompt used in step 1: Decompose \$g\$ into Events \$E\$}} 

\par\vspace{0.5em}

Break down the following paragraph into a list of independent events, listed in chronological order. Resolve all pronouns and referring expressions to their corresponding specific entities. Output only the event list and nothing else.

\par\vspace{0.5em}

\hlc{lightblue}{\{\{\textit{Paragraph}\}\}}

\par\vspace{1em}
\hrule
\par\vspace{1em}

\textbf{\underline{Prompt used in step 3: Incremental Lie Generation}}

\par\vspace{0.5em}

Here is what has happened so far:

\par
\hlc{lightblue}{\{\{\textit{CurrentParagraph}\}\}}

\par
The following new fact occurred after the events described above:

\par
\hlc{lightblue}{\{\{\textit{Event}\}\}}

\par
Please append this new fact directly to the current paragraph. If the addition feels awkward, make only minimal word adjustments to ensure the paragraph flows smoothly—without adding extra narrative details or transitional phrases such as "next" or "following that." Output only the updated paragraph.

\par\vspace{1em}
\hrule
\par\vspace{1em}

\textbf{\underline{Prompt used in rephrasing with different narrative technic}}

\par\vspace{0.5em}

\# Task: Rephrase

\par
Rephrase a given paragraph by applying a different narrative sequencing technique. Follow the steps below carefully:

\par\vspace{0.5em}
\#\# Step 1: Identify the Original Narrative Technique  
Read the original paragraph and determine which of the following sequencing techniques it uses:

\begin{itemize}
  \item \textbf{Chronological Order} — Events are presented strictly in the order they occurred.
  \item \textbf{Flashback} — The paragraph begins with a later or climactic moment, then shifts back to earlier events.
  \item \textbf{Interjection} — The main narrative is interrupted by a relevant insert such as a memory, reflection, or side story.
  \item \textbf{Supplementary Narration} — Contextual background is added to support understanding, even if the details weren't part of the original sequence.
\end{itemize}

\par\vspace{0.5em}
\#\# Step 2: Rephrase Using a Different Technique  
Choose a \textit{different} narrative sequencing method from the list above and rephrase the paragraph accordingly.

\par\vspace{0.5em}
\textbf{Guidelines for Rephrasing:}
\begin{itemize}
  \item Use as much of the original wording as possible.
  \item Do \textbf{not} add any new events or fabricate details not present in the original.
  \item Avoid ambiguous expression.
\end{itemize}

\par
Please output the result in the following format:

\begin{itemize}
  \item \texttt{Original\_Narrative\_Technique: <original\_narrative\_technique>}
  \item \texttt{Choosed\_Narrative\_Technique: <choosed\_narrative\_technique>}
  \item \texttt{Rephrased: <rephrased\_paragraph>}
\end{itemize}

\par\vspace{0.5em}
\hlc{lightblue}{\{\{\textit{Paragraph}\}\}}

\end{minipage}
}
\caption{Prompts used in the data construction process}
\label{tab:prompt-data-construction}
\end{table*}

The prompt used in \textsc{MontageLie} construction is provided in Table~\ref{tab:prompt-data-construction}.

\subsection{Prompts Used in LLM-based Evaluators}
\label{App:PromptEva}
\begin{table*}[ht]
    \centering
    \small
    \noindent\fbox{%
    \begin{minipage}{2.0\columnwidth} 
\tt 
\textbf{\underline{Prompt used in coarse-grained LLM-as-Evaluators} }
\newline
\newline
You will be given a source text. You will then be given one target text to be evaluated. Your task is to rate the information alignment of target text against the source text. Please make sure you read and understand these instructions carefully. Please keep this source text open while reviewing, and refer to it as needed.
\newline
\newline
Evaluation Criteria:
\newline
Consistency (1-5) - the information alignment between the target text and the source text. A consistent target text contains only statements that are entailed by the source source text. Annotators were also asked to penalize target texts that contained hallucinated facts. 1 - worst, 5 - best.
\newline
\newline
Evaluation Steps:
\newline
1. Read the source text carefully and identify the main facts and details it presents.\newline
2. Read the target text and compare it to the source text. Check if the target text contains any factual errors that are not supported by the source text.\newline
4. Assign a score for consistency based on the Evaluation Criteria.
\newline
\newline
Note: only output the score for consistency, no other text.
\newline\newline
Source Text: 
\newline
\hlc{lightblue}{\{\{\textit{Source}\}\}}
\newline\newline
Target Text: 
\newline
\hlc{lightblue}{\{\{\textit{Target}\}\}}
\newline\newline
Evaluation Form (scores ONLY):
\newline
- Consistency:
\newline
\newline
\end{minipage}
}

\caption{The prompts used in benchmarking coarse-grained LLM-as-Evaluators.}
\label{tab:prompt-data-evaluation}
\end{table*}
\begin{table*}[ht]
    \centering
    \small
    \noindent\fbox{%
    \begin{minipage}{2.0\columnwidth} 
\tt 
\textbf{\underline{Prompt used in Decomposer} }
\newline
\newline
Please analyze the following paragraph and extract all independent factual statements, categorized into two types: Event Facts and Descriptive Facts.
\newline
\newline
\#\# Definitions:\newline
- Event Facts:\newline
Time-dependent facts that describe specific actions, changes, occurrences, or emotional/mental states. These involve entities doing something or experiencing something dynamically at a particular point in time, and can be situated along a timeline.\newline
Examples:\newline
The spacecraft entered Mars' orbit after a six-month journey.\newline
Dr. Lin submitted her resignation.\newline
Mary felt happy about her promotion.\newline
\newline
- Descriptive Facts:\newline
Time-independent facts that define, classify, or describe static attributes or relationships of entities. These do not occur at a specific time, and are considered stable or inherent properties.\newline
Examples:\newline
Helianthus is a genus in the daisy family Asteraceae.\newline
Octopuses have three hearts.
\newline\newline
\#\# Instructions:\newline
1. Break down the paragraph into individual, self-contained factual statements.\newline
2. Resolve all pronouns and referring expressions to their full entity names for clarity.\newline
3. Categorize each fact as either an Event Fact or a Descriptive Fact, according to the definitions above.\newline
3. Output two separate lists:\newline
- Event Facts List: List in chronological order.\newline
- Descriptive Facts List: Order does not matter.
\newline\newline
Paragraph: \hlc{lightblue}{\{\{\textit{Paragraph}\}\}}
\newline\newline
Event Facts List and Descriptive Facts List:
\newline
\end{minipage}
}

\caption{The prompts used in decomposer of DoveScore.}
\label{tab:prompt-dovescore-decomposer}
\end{table*}

\begin{table*}[ht]
    \centering
    \small
    \noindent\fbox{%
    \begin{minipage}{2.0\columnwidth} 
\tt 
\textbf{\underline{Prompt used in fact checker} }\newline\newline
Check if the fact is true based on the given context. Return True or False.\newline\newline
Context: \hlc{lightblue}{\{\{\textit{Source}\}\}}\newline\newline
Fact: \hlc{lightblue}{\{\{\textit{Fact}\}\}} True or False?\newline\newline
Output:
\newline
\end{minipage}
}

\caption{The prompts used in fact checker of DoveScore.}
\label{tab:prompt-dovescore-fact-checker}
\end{table*}

\begin{table*}[ht]
    \centering
    \small
    \noindent\fbox{%
    \begin{minipage}{2.0\columnwidth} 
\tt 
\textbf{\underline{Prompt used in sorter} }
\newline\newline
You are a helpful assistant that determines the correct chronological order of events in a paragraph. Do NOT add, remove, or change any events. Only reorder the exact events from the input list.
\newline\newline
Example 1:\newline
\underline{Paragraph}:\newline
Tom woke up early. He brushed his teeth and then had breakfast. After that, he went for a run.\newline
\underline{Events}:\newline
- Tom had breakfast\newline
- Tom woke up\newline
- Tom went for a run\newline
- Tom brushed his teeth
\newline
\underline{Ordered Events}:\newline
[Tom woke up, Tom brushed his teeth, Tom had breakfast, Tom went for a run]\newline\newline
Example 2:\newline
\underline{Paragraph}:\newline
After she went out for lunch, Sarah called her friend. Earlier in the morning, she had replied to a message right after checking her email.
\newline
\underline{Events}:\newline
- Sarah checked her email  \newline
- Sarah went out for lunch  \newline
- Sarah called her friend  \newline
- Sarah replied to a message  
\newline
\underline{Ordered Events}:\newline
[Sarah checked her email, Sarah replied to a message, Sarah went out for lunch, Sarah called her friend] \newline\newline
Now sort the following events based on the paragraph below, and return as a list of events: \newline \newline
Paragraph: \hlc{lightblue}{\{\{\textit{Paragraph}\}\}}\newline \newline
Events: \hlc{lightblue}{\{\{\textit{Events}\}\}}\newline \newline
Ordered Events: 
 \newline 
\end{minipage}
}

\caption{The prompts used in sorter of DoveScore.}
\label{tab:prompt-dovescore-sorter}
\end{table*}
The prompts used in coarse-grained LLM-as-evaluator is provided in Table~\ref{tab:prompt-data-evaluation}.
The prompt used in DoveScore decomposer, fact checker, and sorter are provided in Table~\ref{tab:prompt-dovescore-decomposer}, ~\ref{tab:prompt-dovescore-fact-checker}, and ~\ref{tab:prompt-dovescore-sorter} respectively.

\section{About \textsc{MontageLie}}
\subsection{Shuffle Algorithm Used in Montage-Style Lie Generation}
\label{App:ShuffleAlg}
\SetCommentSty{mycommfont}
\newcommand{\mycommfont}[1]{\textcolor{blue}{\textit{~#1}}}
\SetKwInOut{KwIn}{\textcolor{gray}{Input}}
\SetKwInOut{KwOut}{\textcolor{gray}{Output}}

\begin{algorithm*}[ht]
\caption{\textsc{RandomShuffleWithInversions}}
\label{alg:random-shuffle-inversions}
\KwIn{
    $A = (a_1, a_2, \dots, a_m)$: original list of length $m$\\
    $K$: target inversion count, $0 \leq K \leq \frac{m(m-1)}{2}$
}
\KwOut{
    $B$: a random permutation of $A$ with $\mathrm{Inv}(B) = K$
}

\tcc{Step 1: Initialize maximum inversion sequence}
$m \leftarrow |A|$\;
\For{$i \leftarrow 1$ \KwTo $m$}{
    $e_i \leftarrow m - i$ 
     \tcc*{Maximum possible inversions at position $i$}
}

\tcc{Step 2: Reduce total inversions to desired $K$}
$\Delta \leftarrow \frac{m(m-1)}{2} - K$

\While{$\Delta > 0$}{
    $i \leftarrow \mathrm{UniformRandomInteger}(1, m)$\;
    \If{$e_i > 0$}{
        $e_i \leftarrow e_i - 1$\;
        $\Delta \leftarrow \Delta - 1$
    }
}

\tcc{Step 3: Decode inversion sequence into permutation}
$C \leftarrow$ sorted copy of $A$\;
$B \leftarrow$ empty list of size $m$

\For{$i \leftarrow 1$ \KwTo $m$}{
    $B_i \leftarrow C[e_i + 1]$ 
     \tcc*{Choose the $(e_i+1)$-th smallest available element}
    Remove $C[e_i + 1]$ from $C$
}

\Return{$B$}

\end{algorithm*}
Algorithm~\ref{alg:random-shuffle-inversions} shows that how to obtain a random shuffled list given a list and a target inversion count. 
To generate a random shuffled list with a given number of inversions, the algorithm begins by assigning the maximum possible inversions to each position, assuming the list is fully reversed. It then randomly decreases these inversion values until the total number of inversions equals the target. This creates an inversion sequence that still respects the desired count but adds randomness. Finally, it builds the shuffled list by selecting elements from a sorted version of the original list, guided by the inversion values, ensuring the resulting permutation has exactly the specified number of inversions.

\subsection{Data Distribution}
\label{App:DataDist}
\begin{figure*}[ht]
    \centering
    \includegraphics[width=\textwidth]{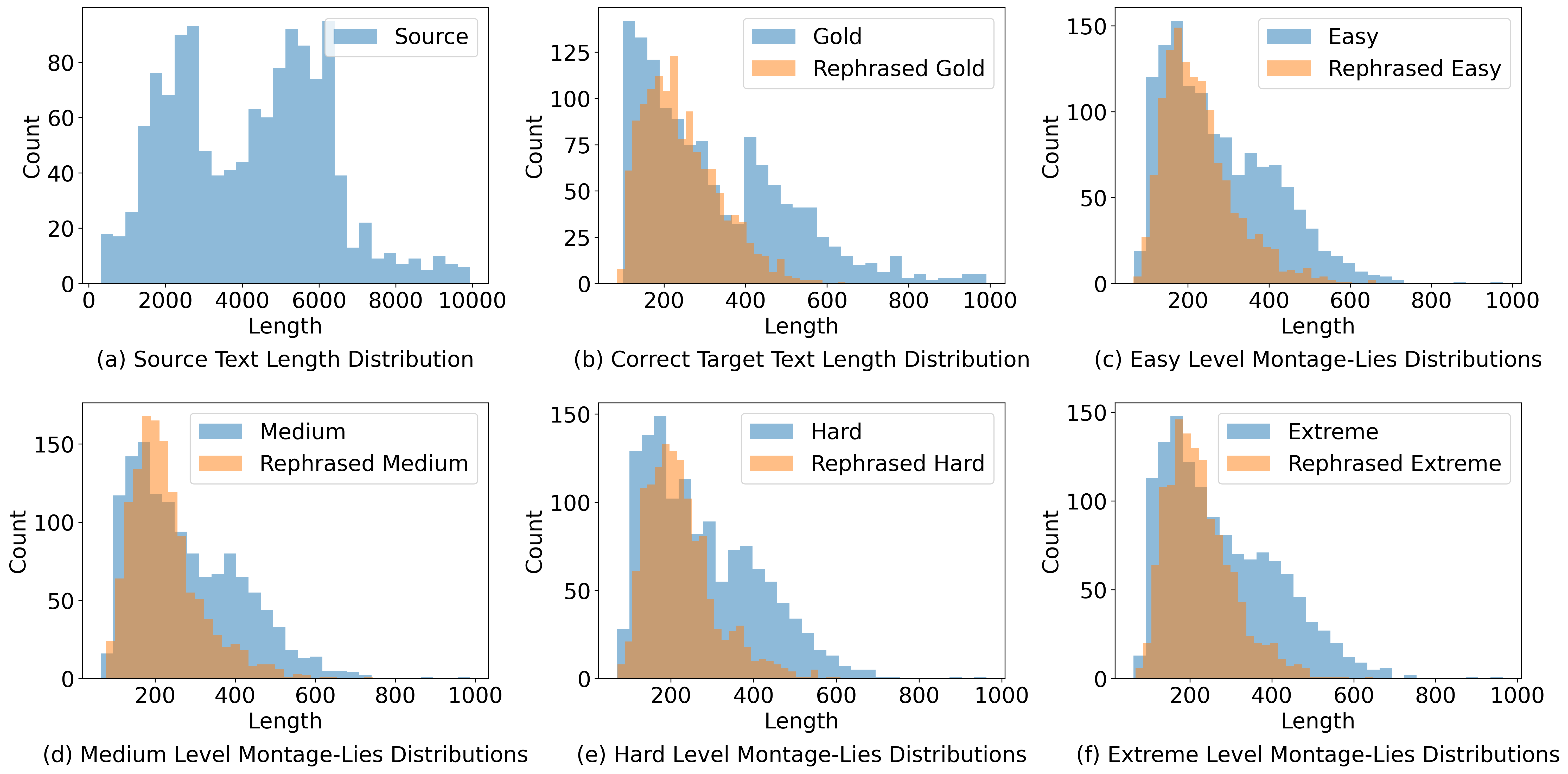}
    \caption{Distribution of words length in MontageLie benchmark.}
    \label{fig:dist-bench-length}
\end{figure*}

\begin{figure}[ht]
    \centering
    \includegraphics[width=.46\textwidth]{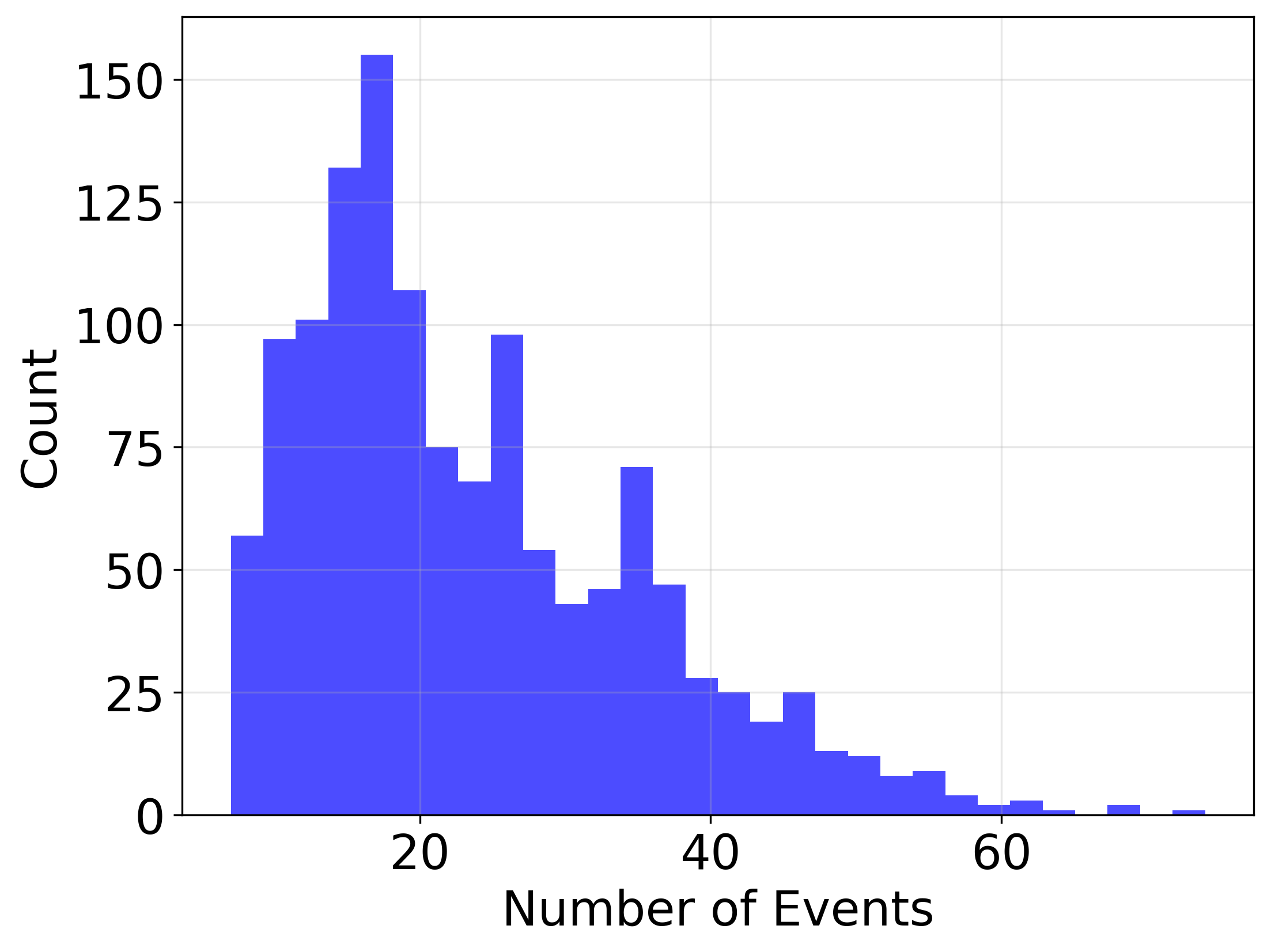}
    \caption{Distribution of number of event decomposed in Step 1 of Montage-Style Lie Generation: Decompose g into Events E.}
    \label{fig:dist-decomposed-event}
\end{figure}

\begin{figure}[ht]
    \centering
    \includegraphics[width=.46\textwidth]{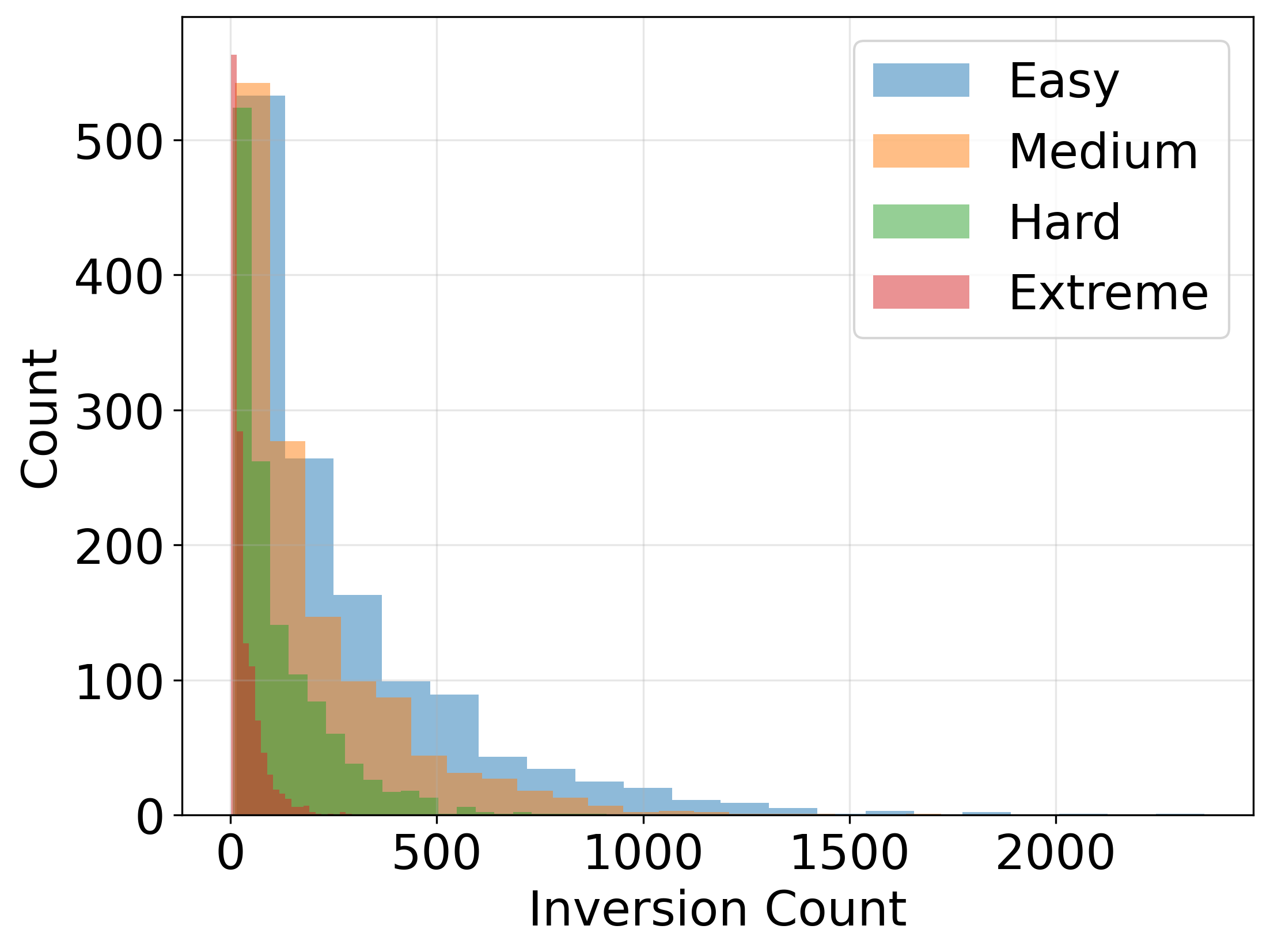}
    \caption{Distribution of Inversion Count Sampled in Step 2 of Montage-Style Lie Generation: Shuffle E with Controlled Difficulty. }
    \label{fig:dist-inversion-count}
\end{figure}

The distribution of the decomposed event number during \textsc{MontageLie} construction, the sampled inversion count when shuffling event lists, and the source and target text length are shown in Figure~\ref{fig:dist-decomposed-event}, ~\ref{fig:dist-inversion-count}, and ~\ref{fig:dist-bench-length}.

\section{Score Distribution on \textsc{montageLie}}
\label{App:ScoreDist}
\begin{figure*}[ht]
    \centering
    \includegraphics[width=\textwidth]{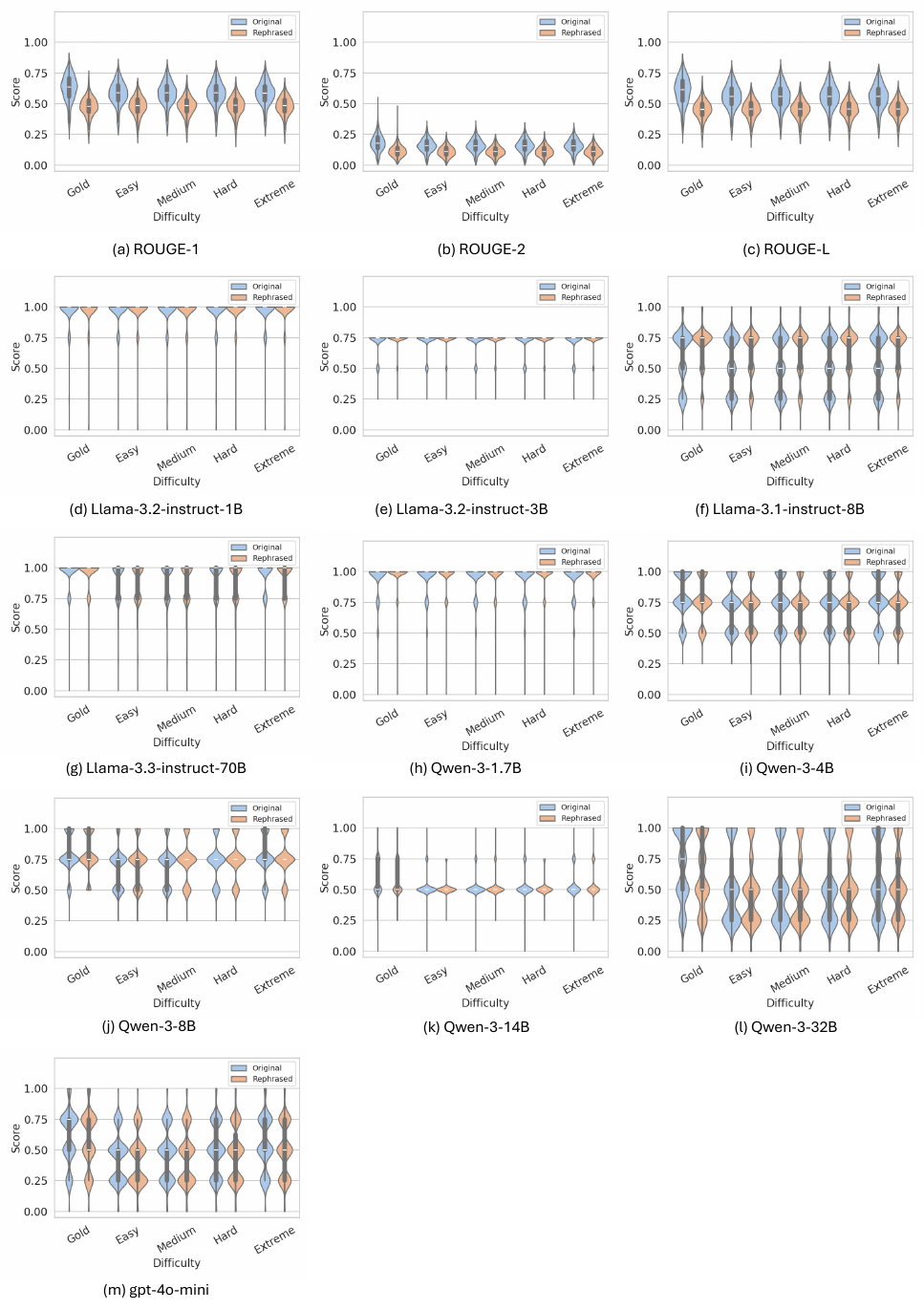}
    \caption{Violin Plots of Score Obtained By Coarse-Grained Evaluators on \textsc{MontageLie}}
    \label{fig:score-dist-coarse}
\end{figure*}

\begin{figure*}[ht]
    \centering
    \includegraphics[width=\textwidth]{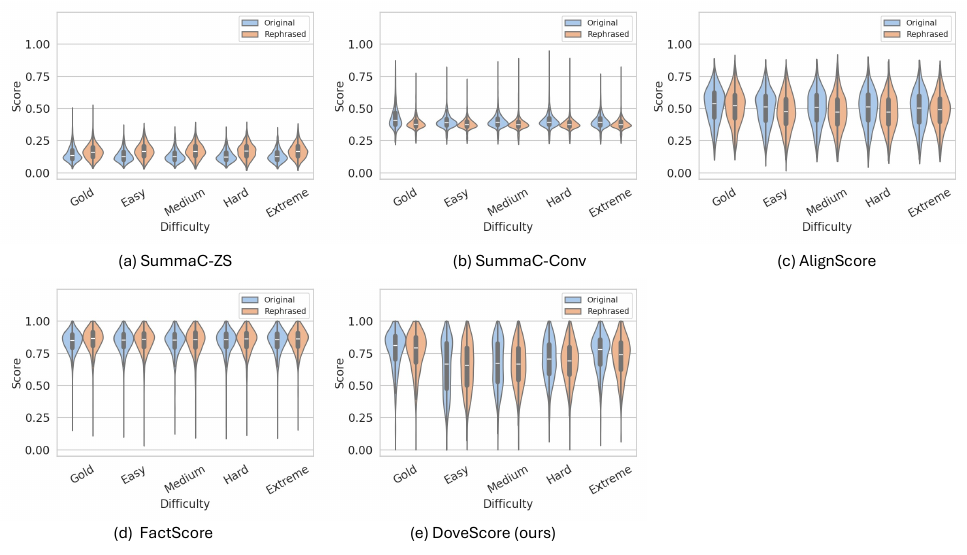}
    \caption{Violin Plots of Score Obtained By Fine-Grained Evaluators on \textsc{MontageLie}}
    \label{fig:score-dist-fine}
\end{figure*}

\begin{figure*}[ht]
    \centering
    \includegraphics[width=\textwidth]{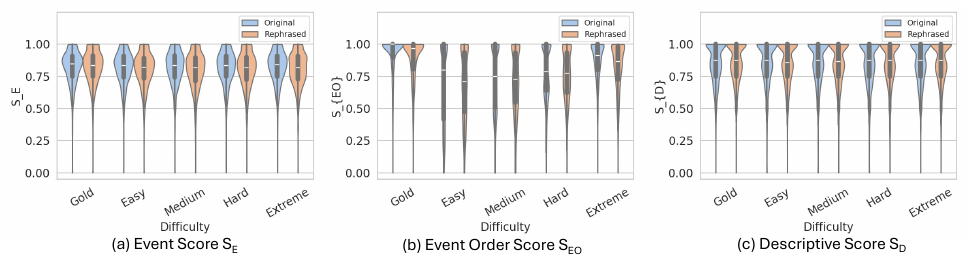}
    \caption{Violin Plots of SubScores obtained by \textsc{DoveScore} on \textsc{MontageLie}. }
    \label{fig:score-dist-subdove}
\end{figure*}

The score distribution of coarse-grained evaluators , fine-grained evaluators are shown in Figure~\ref{fig:score-dist-coarse} and ~\ref{fig:score-dist-fine}. The subscore distribution of DoveScore is provided in Figure~\ref{fig:score-dist-subdove}.


\end{document}